\documentclass{article}

\usepackage[preprint]{nips_2018}

\usepackage[utf8]{inputenc} 
\usepackage[T1]{fontenc}    
\usepackage{algorithm}

\usepackage{hyperref}       
\usepackage{url}            
\usepackage{booktabs}       
\usepackage{amsfonts}       
\usepackage{nicefrac}       
\usepackage{microtype}      

\usepackage{amsmath}
\usepackage{amssymb}

\usepackage{amsmath}
\DeclareMathOperator*{\argmax}{arg\,max}

\usepackage{graphicx}
\usepackage[noend]{algpseudocode}
\usepackage{subfigure}

\title{Autonomous discovery of the goal space to learn a parameterized skill}

\author{
Emilio Cartoni,\\
LOCEN,\\
ISTC-CNR, Rome, Italy\\
\texttt{emilio.cartoni@istc.cnr.it}
\And
Gianluca Baldassarre\\
LOCEN,\\
ISTC-CNR, Rome, Italy\\
\texttt{gianluca.baldassarre@istc.cnr.it}
}

\begin{document}

\maketitle

\begin{abstract}  
A parameterized skill is a mapping from multiple goals/task parameters to the policy parameters to accomplish them.
Existing works in the literature show how a parameterized skill can be learned given a task space that defines all the possible achievable goals.
In this work, we focus on tasks defined in terms of final states (goals), and we face on the challenge where the agent aims to autonomously acquire a parameterized skill to manipulate an initially unknown environment.
In this case, the task space is not known a priori and the agent has to autonomously discover it.
The agent may posit as a task space its whole sensory space (i.e. the space of all possible sensor readings) as the achievable goals will certainly be a subset of this space.
However, the space of achievable goals may be a very tiny subspace in relation to the whole sensory space, thus directly using the sensor space as task space exposes the agent to the curse of dimensionality and makes existing autonomous skill acquisition algorithms inefficient.
In this work we present an algorithm that actively discovers the manifold of the achievable goals within the sensor space.
We validate the algorithm by employing it in multiple different simulated scenarios where the agent actions achieve different types of goals: moving a redundant arm, pushing an object, and changing the color of an object.
\end{abstract}

\section{Introduction} 
A parameterized skill offers a way to accomplish multiple similar tasks by reusing the same skill, more formally, a parameterized skill is a function that maps each task parameters to the parameters of a policy that when executed can achieve it.
Here we focus in particular on tasks that can be defined in terms of goals, i.e. final states that have to be accomplished by the agent within a limited amount of time (`trial').
Existing works in the literature show how such a parameterized skill can be acquired given a task space that defines all the possible achievable goals \citep{Reinhart2016a,Baranes2013,DaSilva2014,DaSilva2014a}. 
However, if we want to develop an autonomous open-ended learning agent, aiming to autonomously acquire an ample repertoire of goals and policies not defined by the experimenter \citep{WengMcClellandPentlandSpornsStockmanSurThelen2001Autonomousmentaldevelopment,ThrunMitchell1995Lifelongrobotlearning,BaldassarreMirolli2013Intrinsicallymotivatedlearninginnaturalandartificialsystems,SantucciBaldassarreMirollisubmGRAILaGoalDiscoveringRoboticArchitectureforIntrinsicallyMotivatedLearning}, it may be difficult to define the task space of the achievable goals at design time. 
%

One easy way to define a task space which will encompass all possible goals is to define the goal space as being the sensor space defined by the agent sensors.
As an example, we may define the goal space as all the possible readings from the robot cameras (sensor space).
In this way, anything that the robot can perceive might be considered as a possible goal.
This will ensure that all achievable goals, as long as the robot can see them, form a subset of such sensor space.
However, the sensor space may be huge in comparison to the actual (achievable) goal space: a 320x240 pixel camera image has more than 80000 dimensions and most of the possible camera readings that one may construct from those pixels are not images that the agent can actually accomplish with its actions.
As we will show, current algorithms for learning parameterized skills do not work well in such highly dimensional spaces where the achievable goals form a small subset of the whole sensor space (see Sec.~\ref{sec:one}). 

We thus propose here a new algorithm that actively learns the manifold of the achievable goals embedded in the sensor space by building a graph representation of the achieved goals.
As it focuses on the manifold of achievable goals, our algorithm escapes the curse of dimensionality 
and outperforms current approaches, 
making it possible to use a ``generic' a priori goal space defined on the basis of the sensors space such as the image space. 
We validate the algorithm by employing it in multiple different simulated scenarios where the agent actions achieve different types of goals (moving an arm to desired postures, pushing an object to desired locations, changing the color of an object to specific colors). 

    
\section{Open-ended autonomous learning of parameterized skills}\label{sec:one}

\subsection{The objective}

Let us consider an agent with sensors readings that lie on a sensor space $S \in \mathbb{R}^n$.
The agent interacts with an environment through a policy $\pi \in \Pi$ parameterized with $\psi \in \Psi$ and terminating after a certain amount of time has elapsed.
The world state resulting from the performance of the policy will be a perceived outcome $o = envi(\pi(\psi))$, with $o \in O \subseteq S $ and $|O|<<|S|$  (note we say ``outcomes'' to refer to action consequences, and ``goals'' to refer to ``desired states'', but the two belong to the same space $O$).
The objective of the agent is to learn the parameterized skill $\Theta$ 
that maps every outcome $o$ to a policy parameter set, $\Theta:O \rightarrow \Psi$, 
so that we can later ask the agent to achieve any goal $g \in O$.
In particular, we test the agent's performance $perf$, probing the quality of the parameterized skill $\Theta$, by asking it to reach $N$ goals randomly sampled from $O$:
\begin{equation}
  perf(\Theta) = \sum\limits_{i=1}^N dist(g_i,env(\pi(\Theta(g_i)))) / N
\end{equation}
where $dist$ is a function that returns 1 if the achieved outcome $o = env(\pi(\Theta(g)))$ is sufficiently similar to the desired goal $g$ and 0 otherwise.

\subsection{Previous approaches}
Current approaches such as \cite{Reinhart2016a,Baranes2013,DaSilva2014,DaSilva2014a,Pere2018} assume knowing the outcome set $O$.
However, if we do not know $O$ beforehand, we might have to substitute it with the sensor space $S$ instead.
If not all sensor readings are achievable as outcomes, $S$ is going to be a much larger set than $O$ and sampling from $S$ would produce many goals that the agent cannot achieve.
This would make an approach such as \cite{DaSilva2014} unfeasible since it assumes that all drawn tasks are solvable by policy optimization. Even adding some mechanism that ``gives up'' policy optimization after a few trials would not help if most goals drawn from $S$ are not achievable, as in the case $S$ represents images from a robot camera: the algorithm would spend most of its time pursuing goals represented by images formed by random pixels.
Both \cite{DaSilva2014a} and \cite{Baranes2013} do not sample $O$ uniformly but use mechanisms to bias the sampling towards the most promising regions first, those where the algorithm expects to make most progress. In theory this could help to avoid the non achievable part of $S$, but it would still need to sample some achievable goals first which is practically impossible with large spaces such as the camera image space.

An interesting approach is proposed in \cite{Pere2018} where the agent does not learn on a predefined goal space but tries to first learn a latent space from raw sensor space observations and then use this latent space as its outcome.
However, in \cite{Pere2018} the raw sensor space observations are provided to the agent by an unspecified external process which samples all possible outcomes (possibly also some unachievable ones) and feeds the corresponding sensor readings to the agent.
So while the agent is not provided with an engineered goal space, the knowledge on such space is still needed to make the approach to work.

In contrast with these approaches, \cite{Reinhart2016a}'s \textit{skill babbling} does not require sampling directly from the whole $O$.
\textit{Skill babbling} instead starts from a single $(policy, outcome)$ pair and gradually explores the outcome space by perturbing the known outcomes and their corresponding policies.
A similar approach to discover new outcomes was also used in \cite{Seepanomwan2017}, although in that case the focus was on the discovery of discrete outcomes and policies rather than on learning a continuous goal-policy parameterized skill.
However, as we will see in the following experiments, the exploration strategy proposed by \cite{Reinhart2016a} has limitations when dealing with camera images due to the inability of exploring such a high-dimensional sensor space.
Due to the similarity to our approach, which also starts from a seed and then gradually builds up the goal/policy repertoire, we will use \textit{skill babbling} as a baseline to evaluate our algorithm.

\subsection{The algorithm: Active Goal Manifold Exploration (AGME)}

To avoid the problems due to using $S$ instead of $O$, we propose an algorithm that gradually builds a repertoire of known (achieved) outcomes $O_A$ by trying to progressively discover the manifold $O$ inside $S$ (see Algorithm~\ref{alg:euclid}).
The algorithm keeps track of all achieved outcomes $O_A$ and at each trial chooses one of them as a \textit{basis goal} to discover new outcomes.
The algorithm makes the assumption that goals near each other have similar policies, so it tries to discover new goals by perturbing the policy corresponding to the basis goal. 
The choice of the basis goal for each trial is based on the idea that the system should favor goals that are more likely to have undiscovered outcomes near them.
In particular, the algorithm builds a k-neighbor graph on $O_A$ (previously achieved outcomes) and then measures the average distance of each outcome of $O_A$ from its k-neighbors.
The outcome that has the maximum average distance from its k-neighbors is the basis goal.
The reason for choosing this basis goal is the assumption that an outcome which is farthest from its discovered neighbors has the highest potential to have undiscovered neighbors.
On the contrary, an outcome whose discovered neighbors are close is probably in an area of the goal space which is already well explored.
The algorithm then generates and executes a new policy which is the same as the one that achieved the chosen basis goal plus some Gaussian noise.

AGME builds a repertoire of outcomes, with their corresponding policies, but it does not explicitly create or learn a parameterized skill $\Theta$.
Different strategies can be employed to construct $\Theta$ from the repertoire of goals and policies.
Since this is not the focus of this work, in the following sections we simply assume that the agent uses a k-neighbor regression with $k=1$: i.e. when asked to achieve certain goal, the parameterized skill $\Theta(g)$ simply returns the policy of the most similar discovered outcome from $O_A$.

\begin{algorithm}
\caption{Active Goal Manifold Exploration}\label{alg:euclid}
\begin{algorithmic}[1]
\Require Set of achieved outcomes, $O_A$, and their policy parameters $\Phi$
\While{$True$}
\ForAll{$o_i$ in $O_A$} 
\State Compute k-neighbors of $o_i$
\State $\delta_{o_i}\gets \sum\limits_{j=1}^k d(o_i,o_j) / k$
\Comment{Compute average distance from neighbors}
\EndFor
\State $basis \gets \argmax\limits_i \delta_{o_i}$ \Comment{Choose farthest outcome}
\State $ \boldmath{\epsilon} \sim \mathcal{N}(\boldmath{0},\,\sigma^{2}) $ \Comment Exploration noise vector
\State $\phi_{new} \gets \phi_{basis} + \epsilon $ \Comment Generate new policy parameters
\State $o_{new} \gets env(\pi(\phi_{new})) $\Comment Execute policy in the environment, get a new outcome
\State $O \gets O \cup o_{new}$ \Comment Add outcome to the repertoire
\State $\Phi \gets \Phi \cup \phi_{new}$ \Comment Add policy to the repertoire
\EndWhile

\end{algorithmic}
\end{algorithm}

\subsection{Experiment 1: learning an arm inverse model}
\label{sec:Exp1}

\subsubsection{Setup}
In the first experiment, the agent learns to move the end-point of a planar 3-link arm everywhere in its working space.
The input to the system is either formed by the x,y position of the arm end-point (which might be produced by a suitable sensory preprocessing), or by the raw 50x50 pixel RGB image of the arm coming from a simulated camera (Fig.\ref{fig:arm}).
On each trial, the agent policy moves the arm by specifying the three arm joint angles constrained within -60 and +60 degrees.

\subsubsection{Results}
\begin{figure}[!htb]
\minipage{0.32\textwidth}
  \includegraphics[width=\linewidth]{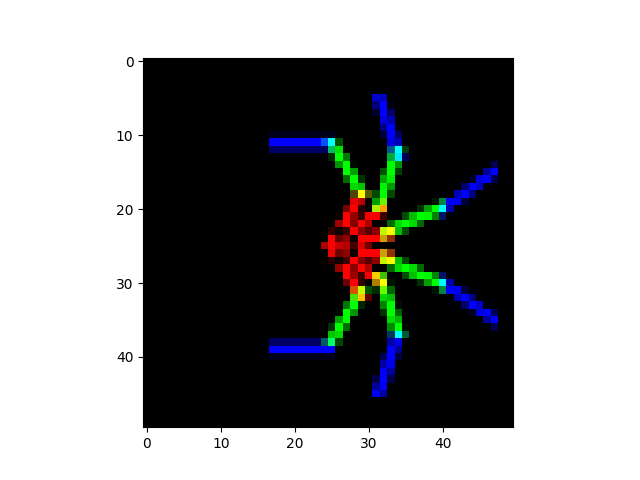}
\caption{Experiment 1: the agent has a 3-link planar arm, centered on a plane imaged by a 50x50 pixel camera. The images shows 6 possible positions of the arm superimposed.}\label{fig:arm}
\endminipage\hfill
\minipage{0.64\textwidth}
  \includegraphics[width=\linewidth]{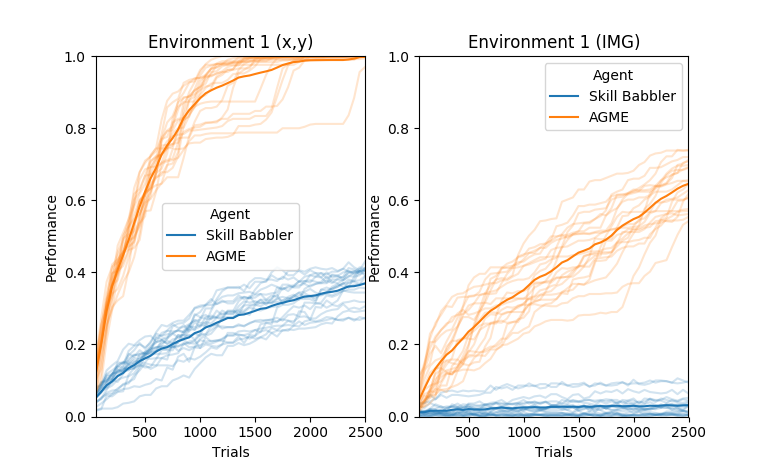}
\caption{Results of Experiment 1. Light colored lines are individual simulations, while darker lines show their average. 
Left: AGME has a steady progress even when the outcome space has to be discovered within the sensor space (images).
Right: \textit{skill babbling} fails to discover the goal manifold in the image space.}
\label{fig:exp1_res}
\endminipage
\end{figure}

\begin{figure}[!htb]
\includegraphics[width=\textwidth]{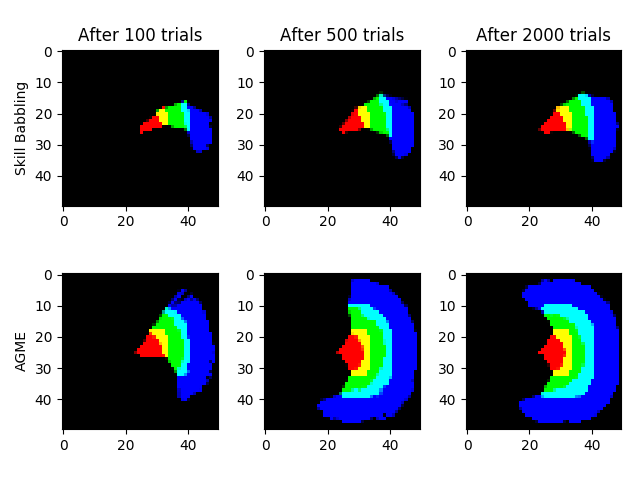}
\caption{
Superposition of the outcomes (arm positions) discovered by \textit{skill babbling} and AGME when using the image space. 
AGME discovers more arm positions than \textit{skill babbling} which remains mostly confined around the starting seed.}
\label{fig:exp1_res2}
\end{figure}

Fig.~\ref{fig:exp1_res} shows the agent learning when the outcome is encoded either as the x,y arm end-point position or as the image of the arm.
The AGME algorithm performs well both with the predefined outcome space and with the image space.
Instead, \textit{skill babbling} has a limited progress when faces the image space.
Fig.~\ref{fig:exp1_res2} shows how AGME progressively discovers the whole goal manifold $O$ embedded in $S$ (i.e., all possible images corresponding to all the arm postures) while \textit{skill babbling} gets stuck.

\subsection{Experiment 2a: pushing an object to desired locations}\label{sec:Exp2a}

\subsubsection{Setup}
\begin{figure}[!htb]
\centering
\subfigure[Start of trial]{
\includegraphics[width=.325\textwidth]{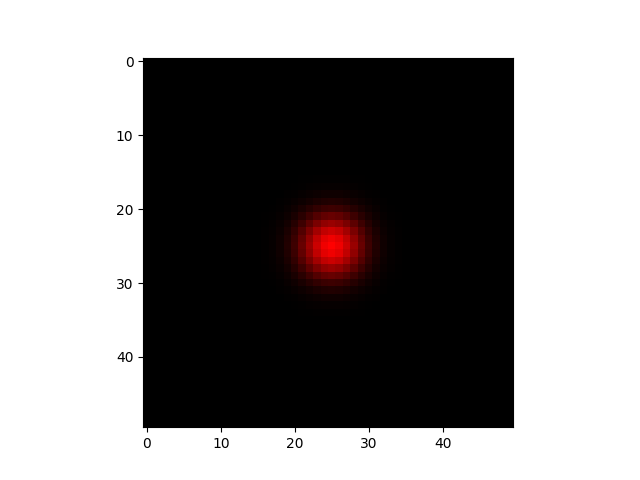}
}
\subfigure[During the trial]{
\includegraphics[width=.255\textwidth]{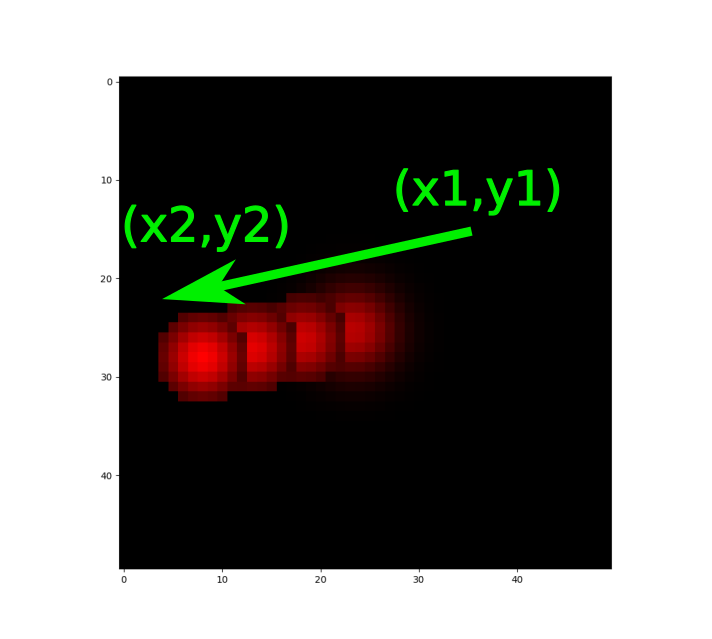}
}
\subfigure[End of trial]{
\includegraphics[width=.325\textwidth]{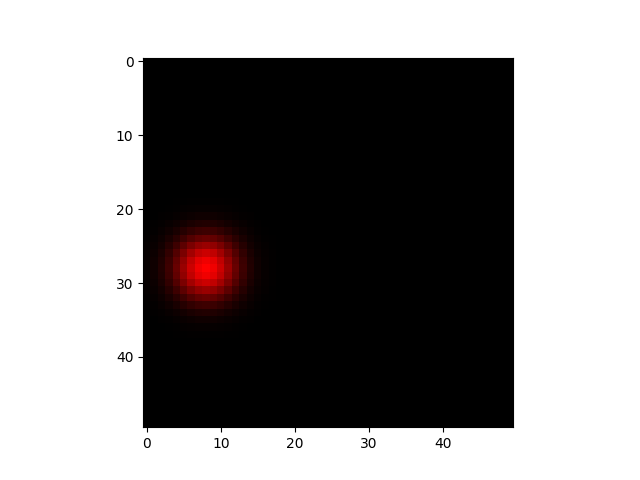}
}
\caption{Experiment 2a: a red circular object is placed at the center of a 2D working space.
The agent can move the object by performing a linear trajectory with its arm, for example following the 
policy $<x1,y1,x2,y2>$ indicated by the green segment.
If the trajectory intersects the object, the object is moved to its final end point, e.g. here to  $<x2,y2>$.}
\label{fig:moveTask}
\end{figure}

In the second experiment the agent's objective is to learn to push an object to a desired position.
The environment is formed by a circular object lying on a 2D square workspace. 
The plane is observed by the agent with a fixed 50x50 camera.
The policy performed at each trial specifies a linear trajectory (\textit{start point} and \textit{end point}) for a planar-arm end-point.
At the beginning of each trial the object is set to the central position of the working space.
If the arm end-point crosses the object, the object will be displaced to the end point specified by the policy (see Fig.~\ref{fig:moveTask}).
The agent can observe the outcome of the performed policies either as the final x,y position of the object or as the corresponding camera image.

\subsubsection{Results}
\begin{figure}[!htb]
\includegraphics[width=.7\textwidth]{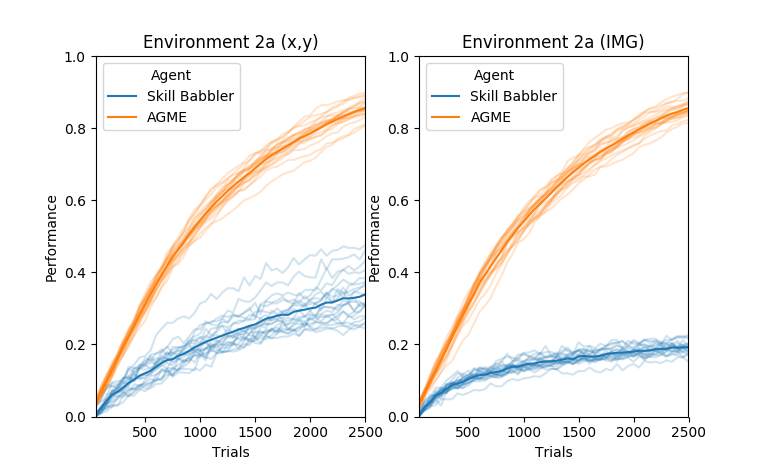}
\caption{
Results of Experiment 2a. 
Light colored lines are individual simulations, while darker lines show their average. 
AGME rapidly reaches maximum performance both if the goal space (object x,y position) or the camera sensor space are provided. 
Instead, \textit{skill babbling} plateaus at a lower level when images are used.}
\label{fig:exp2a_res}
\end{figure}

\begin{figure}[!htb]
\includegraphics[width=.7\textwidth]{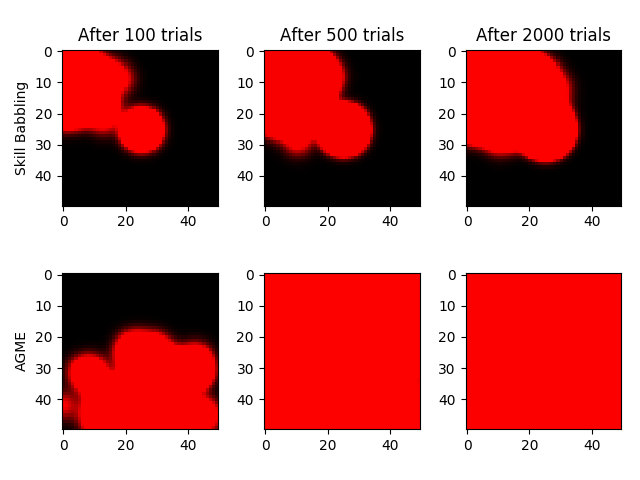}
\caption{
The images show a superposition of all the outcomes (object positions) experienced by \textit{skill babbling} and AGME after 100, 500 and 2000 trials.
Again AGME is able to discover the goal space embedded in the sensory space, thus becoming able to move the object to all parts of the working space.
Instead, \textit{skill babbling} is limited.}
\label{fig:exp2a_res2}
\end{figure}

The results show the agents learning when the outcomes are predefined as the x,y positions of the object and as images.
As in the previous experiment, AGME is able to learn in both the x,y and image scenarios, while \textit{skill babbling} cannot cope with images (Fig.~\ref{fig:exp2a_res} and Fig.~\ref{fig:exp2a_res2}).

\subsection{Experiment 2b: pushing an object everywhere using Dynamic Movement Primitives}
\label{sec:Exp2b}

This experiment uses the same setup as the previous one, except in this case the policy is a dynamic movement primitive (DMP), starting from a fixed point.
\subsubsection{Setup}
The environment is the same as in the previous experiment (Sec.\ref{sec:Exp2a}).
However, this time the agent policy is a dynamic movement primitive: policy parameters specify 10 weights (5 for the x-axis and 5 for the y-axis) of the DMP Gaussians, plus the DMP end point. The start point of the DMP is fixed, as shown in Fig. \ref{fig:exp2b_env}.
\subsubsection{Results}
\begin{figure}[!htb]
\centering
\subfigure[Environment]{
\includegraphics[width=.35\textwidth]{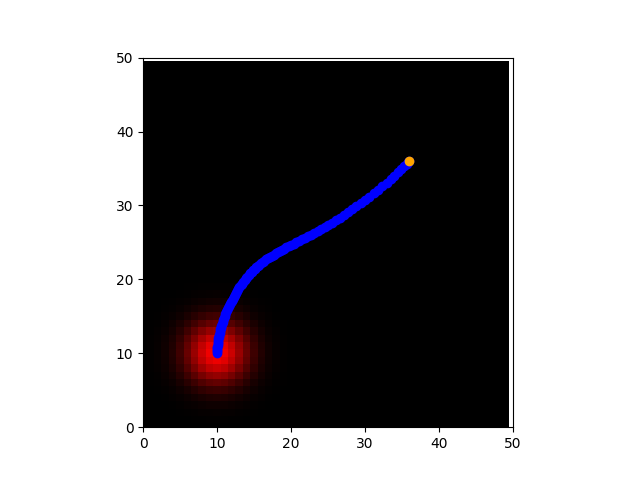}\label{fig:exp2b_env}
}
\subfigure[Results]{
\includegraphics[width=.6\textwidth]{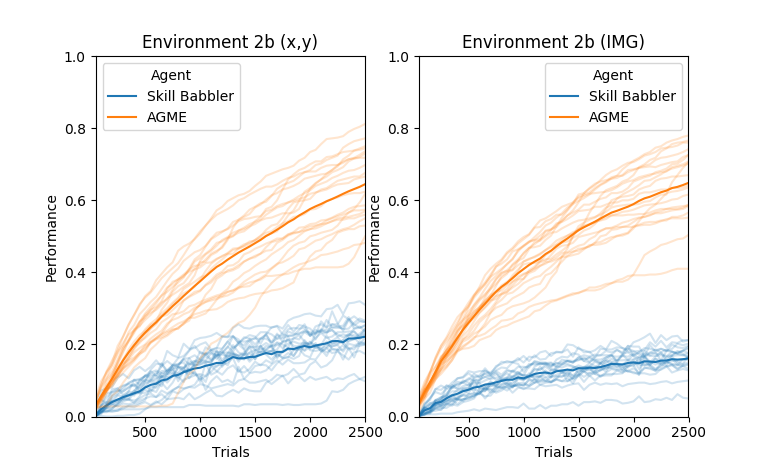}\label{fig:exp2b_res}
}
\caption{\textbf{Experiment 2b:} \ref{fig:exp2b_env} the environment is still the same as Experiment 2a, but this time the policy is expressed as a DMP, which permits curved movements. A sample DMP is depicted in blue. The fixed starting point for all the DMPs is displayed in orange.
In \ref{fig:exp2b_res} light colored lines show the performance of individual simulations, while darker lines show their average. As in the previous experiment AGME does not seem to suffer from switching from x,y outcome space to camera images, while \textit{skill babbling} is slowed.}
\end{figure}

Fig. \ref{fig:exp2b_res} shows that using a DMP as the agent policy yields similar results as the previous experiments.

\begin{figure}[!htb]
\includegraphics[width=.7\textwidth]{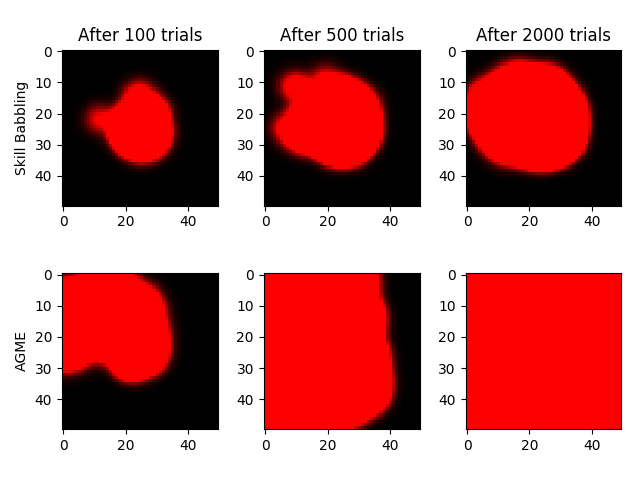}
\caption{The images show a superposition of all the outcomes (object positions) experienced by the \textit{skill babbling} and the AGME algorithm after 100, 500 and 2000 trials. Again AGME rapidly explores the environment and through the DMP moving the object to all parts of the space. Notice that to move the object in the upper right corner, the algorithm had to discover a DMP trajectory that first goes to the center and then moves back up. The \textit{skill babbling} algorithm exploring is instead very limited.}
\end{figure}

\subsection{Experiment 3 - Touching different parts to achieve different colors}\label{sec:Exp3}
In the third environment the agent learns to touch different parts of an object to make it change its appearance to different colors (see Fig.\ref{fig:exp3_env}). 
As in previous experiments, learning is simulated both using a pre-defined outcome and using camera images.
\subsubsection{Setup}
The environment is constituted by a single circular red object lying at the center of a square.
At each trial the agent policy specifies the linear trajectory (start point and end point) of an effector.
If the effector trajectory intersect the circle, the circle will change its color from red to another color, which depends on which part of the circle was hit (see \ref{fig:exp3_env}).
The outcome is passed to the agent is either the point on which the circle was hit or an image.
At the start of each trial, the object color is reset to red.
\subsubsection{Results}
\begin{figure}[!htb]
\centering
\subfigure[Environment]
{
\begin{tabular}{cc}
\includegraphics[width=.18\textwidth]{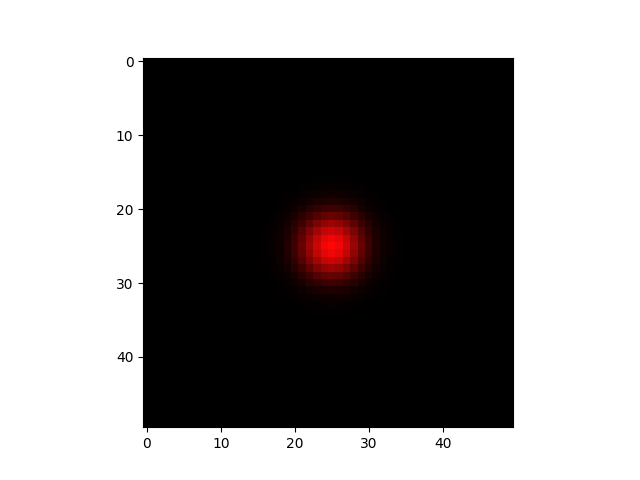} &
\includegraphics[width=.18\textwidth]{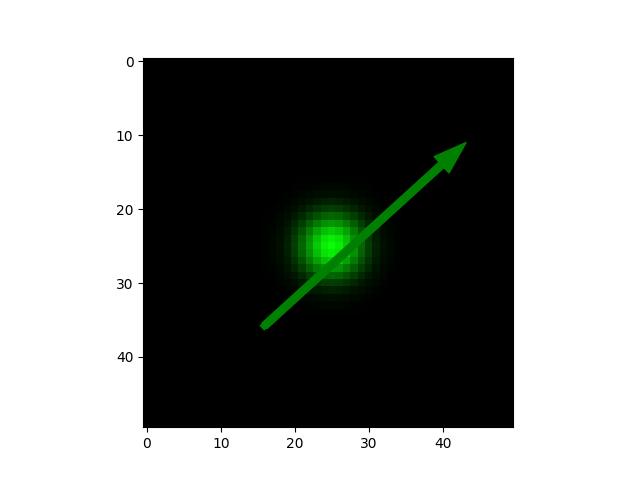}
\\
\includegraphics[width=.18\textwidth]{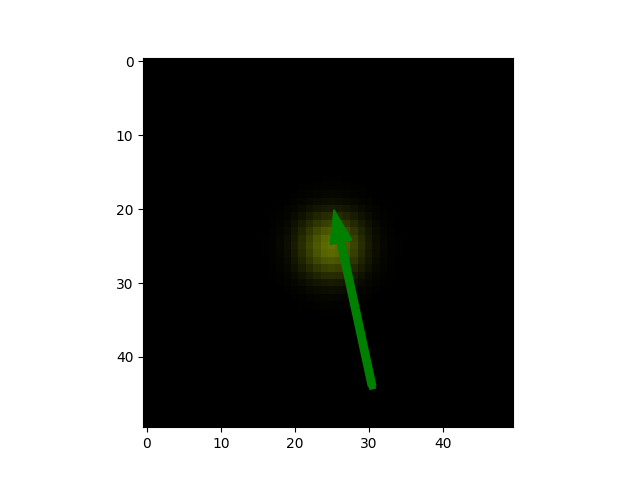} &
\includegraphics[width=.18\textwidth]{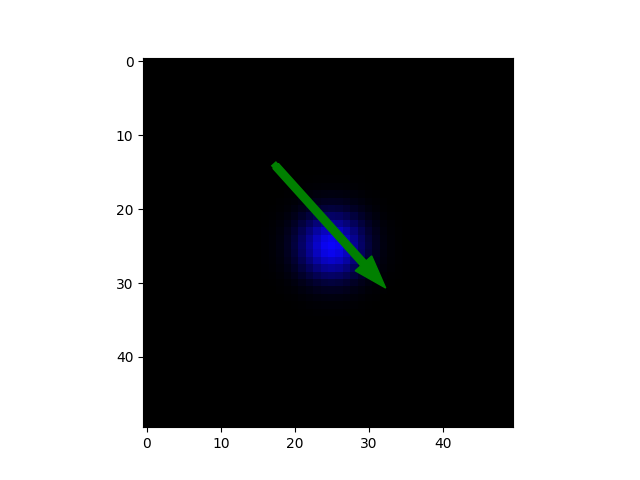}\label{fig:exp3_env}
\end{tabular}
}
\subfigure[Results]{
\begin{tabular}{cc}
\includegraphics[width=.5\textwidth]{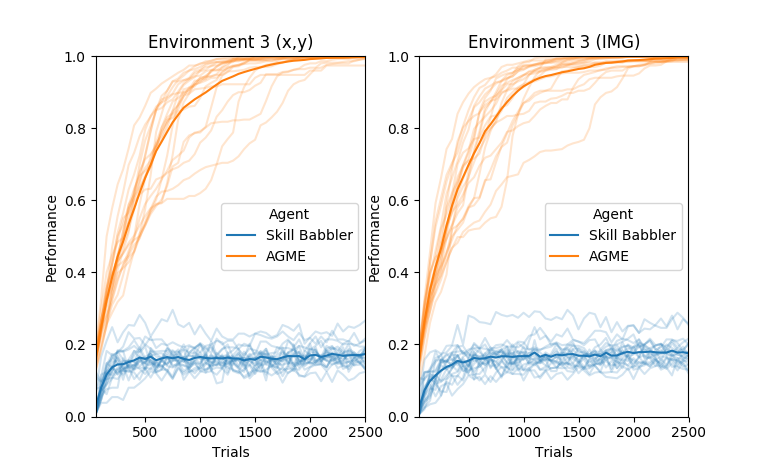}\label{fig:exp3_res}
\end{tabular}
}
\caption{\textbf{Experiment 3:} in this environment, hitting the object does not move it, but it changes its color depending on the point where it is hit (\ref{fig:exp3_env}). As in previous experiment, AGME fares well using both types of outcomes (\ref{fig:exp3_res}). Light colored lines are individual simulations, while darker lines show their average.}
\end{figure}

Fig.\ref{fig:exp3_res} shows the results: again the AGME algorithm works both with pre-defined outcome variables and with images, while \textit{skill babbling} works only with the former.

\section{Discussion}

The experiments show that our AGME algorithm adapts well to different environments involving different types of outcomes and policies.
In all these environments, using the sensor space (images from the camera) works well, often with performance similar to when using a predefined goal space tailored to the specific tasks to solve.
Compared to existing approaches, our algorithm escapes the problem of not knowing the outcome space in advance thanks to two different features:
1) it never samples from the outcome space, so it does not require to know it in advance
2) it does not need to sample from the sensory space either, so it avoids the problem that in such space valid goals are very rare.
Instead, AGME gradually discovers valid goals exploring the sensory space from the known ones and represents them as a graph.
Based on this representation, it measures the distances between discovered goals to actively explore areas of the sensory space having a higher chance to host new goals.
The \textit{skill babbling} algorithm \citep{Reinhart2016a} also builds a gradual representation of the goal space.
However, its exploration of the goal space in the simulated environments was poor and in particular it did not manage to cope with the image space.
Indeed, its exploration mechanism based on generating goals within ``bubbles'' around known goals does not actively drive exploration towards more promising space areas but basically relies on noise.
Also, \textit{skill babbling} uses a mechanism to generate a new policy that adds noise to known goals and then computes the possible corresponding policy through regression, thus it relies on having a good regressor mapping goals to policies.
Adding noise to an image often does not result in a valid goal, plus the regressor may also introduce further problems to the generated policy if the regressor is not yet sufficiently accurate. 
Our non-parametric approach instead does not require a regressor and for the final performance it uses the policies of the goals that are most similar to the requested goals.
To generate the policy, here we in particular used a simple k-neighbour regressor but more sophisticated regressors trained with the collected goals/policies might be used to achieve a better generalization to non experienced goals (this is for example done in \cite{Forestier2017}).

\section{Conclusion}

Current approaches for parameterized skill learning require to know the goal space in advance.
Instead, our AGME algorithm can be used with a very general and broad goal space corresponding to the raw sensory space, such a camera-image space.
These spaces possibly have a very high number of dimensions and goals that are achievable within them represent only a tiny fraction of the whole space.
AGME gradually builds a representation of the goal space based on the goals it progressively discovers and based on this representation tries to find the most promising areas to further discover new goals thus identifying the whole manifold of achievable goals within the larger sensor space.
The results of the proposed experiments show how AGME can work with different types of goals and policies, thus showing generality across different scenarios.
This ability to work in different environments without the need for predefined variables describing the task space is a fundamental step towards real open-ended autonomous learning systems.

\section*{Acknowledgments}
This research has received funding from the European Union's Horizon 2020 Research and Innovation Program under Grant Agreement No 713010 (GOAL-Robots -- Goal-based Open-ended Autonomous Learning Robots).

\bibliography{biblio}
\bibliographystyle{apalike} 

\end{document}